\documentclass[10pt,twocolumn,letterpaper]{article}

\usepackage{cvpr}              

\usepackage[dvipsnames]{xcolor}
\newcommand{\red}[1]{{\color{red}#1}}
\newcommand{\green}[1]{{\color{green}#1}}

\usepackage[ruled,linesnumbered,titlenumbered]{algorithm2e}
\SetKwInOut{Require}{Require}
\SetKwFor{For}{for}{do}{end~for}
\SetKwComment{Comment}{$\triangleright$\ }{}

\usepackage[export]{adjustbox}

\graphicspath{ {./images/} }

\definecolor{cvprblue}{rgb}{0.21,0.49,0.74}
\usepackage[pagebackref,breaklinks,colorlinks,citecolor=cvprblue]{hyperref}

\usepackage{amsmath,amsfonts,bm}

\def\eqref#1{equation~\ref{#1}}

\def\1{\bm{1}}

\DeclareMathAlphabet{\mathsfit}{\encodingdefault}{\sfdefault}{m}{sl}
\SetMathAlphabet{\mathsfit}{bold}{\encodingdefault}{\sfdefault}{bx}{n}

\newcommand{\bc}{\mathbf{c}}

\newcommand{\bx}{\mathbf{x}}
\newcommand{\bxref}{\mathbf{x}_\mathrm{ref}}

\newcommand{\btheta}{{\boldsymbol{\theta}}}

\newcommand{\bepsilon}{{\boldsymbol{\epsilon}}}

\usepackage{mathtools}
\usepackage{comment}
\usepackage{physics}

\def\paperID{}
\def\confName{CVPR}
\def\confYear{2024}

\title{ScribbleGen: Generative Data Augmentation Improves\\ Scribble-supervised Semantic Segmentation}

\author{
{Jacob Schnell$^1$ \thanks{Work done during an internship at UC Merced. \\ Published at workshop on SyntaGen - Harnessing Generative Models for Synthetic Visual Datasets at IEEE Conference on Computer Vision and Pattern Recognition (CVPR), 2024.}
\qquad
Jieke Wang$^2$
\qquad
Lu Qi$^2$
\qquad
Vincent Tao Hu$^3$
\qquad
Meng Tang$^2$}\\
{$^1$University of Waterloo
\qquad
$^2$University of California, Merced
\qquad
$^3$CompVis Group, LMU Munich }
}

\begin{document}
\maketitle
\begin{abstract}
Recent advances in generative models, such as diffusion models, have made generating high-quality synthetic images widely accessible. 
Prior works have shown that training on synthetic images improves many perception tasks, such as image classification, object detection, and semantic segmentation. We are the first to explore generative data augmentations for scribble-supervised semantic segmentation.
We propose ScribbleGen, a generative data augmentation method that leverages a ControlNet diffusion model conditioned on semantic scribbles to produce high-quality training data. 
However, naive implementations of generative data augmentations may inadvertently harm the performance of the downstream segmentor rather than improve it.
We leverage classifier-free diffusion guidance to enforce class consistency and introduce encode ratios to trade off data diversity for data realism.
Using the guidance scale and encode ratio, we can generate a spectrum of high-quality training images. 
We propose multiple augmentation schemes and find that these schemes significantly impact model performance, especially in the low-data regime.
Our framework further reduces the gap between the performance of scribble-supervised segmentation and that of fully-supervised segmentation.
We also show that our framework significantly improves segmentation performance on small datasets, even surpassing fully-supervised segmentation. The code is available at \url{https://github.com/mengtang-lab/scribblegen}.

\begin{figure}
    \centering
    \includegraphics[width=0.5\textwidth]{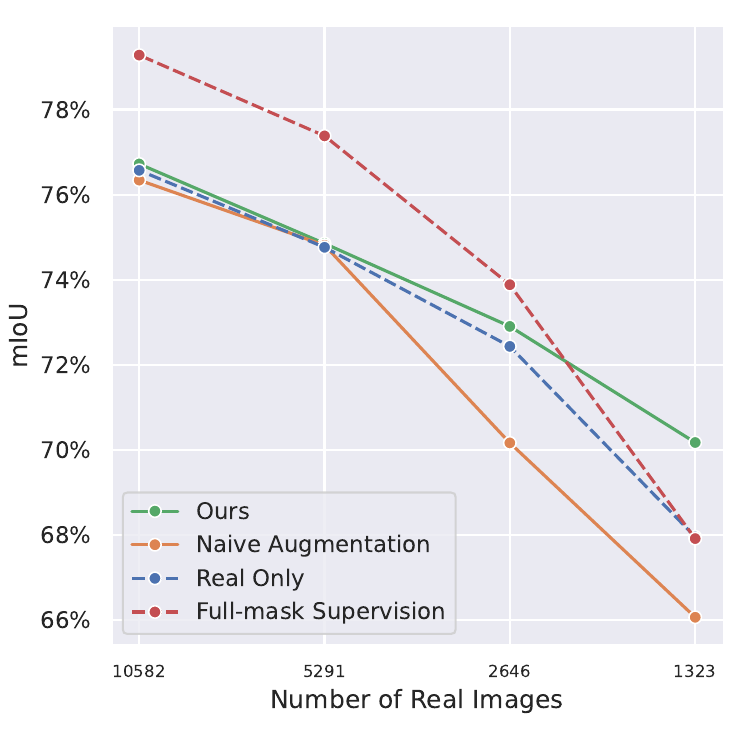}
    \caption{
    Segmentation model performance on PascalVOC and its subsets. 
    All results other than full-mask supervision use scribble-supervision with ResNet-based RLoss~\cite{rloss} model. Naive data augmentation (i.e., fixed encode ratio $\lambda=1.0$) harms model performance, especially in the low-data regime, while our augmentation scheme (Adaptive $\lambda$ Sampling) improves performance.
    }
    \label{fig:ssl}
\end{figure}

\end{abstract}    
\section{Introduction}
\label{sec:intro}

With the massive leaps forward in modern deep learning, machine learning model capacity has never been higher, with some models reaching billions of parameters \cite{dehghani2023scaling, vit}.
However, for many tasks, the size and complexity of datasets have not kept up with the explosion in model capacity.
Since machine learning models perform with a large and rich training dataset, the question of scaling datasets to match model sizes is increasingly pressing.
For tasks like fully-supervised semantic segmentation (FSSS), however, this is especially expensive due to the need for dense pixel-level annotations.
These annotations must often also be produced by experts with domain-specific knowledge, exacerbating the costs of data labeling even further.

Weakly-supervised semantic segmentation (WSSS) seeks to reduce the requirement for dense annotations by using weak annotations.
Such methods include scribble-supervised semantic segmentation, where only a fraction of pixels along some lines (scribbles) are provided. However, these methods still lag behind fully-supervised alternatives regarding segmentation quality, with state-of-the-art methods still achieving 2-4\% lower mIoU \cite{AGMM, TEL} relative to fully-supervised models.

Another strategy is to produce synthetic training data using image-generative models. Prior works have shown that using Generative Adversarial Networks (GANs) to produce training data improves results in image classification \cite{syntheticganclass} and semantic segmentation \cite{Datasetgan,ganaug}, among other tasks.
Diffusion models~\cite{sohl2015deep,ddpm,song2021scorebased_sde}, a well-known type of generative models, have demonstrated strong performance in terms of controllability~\cite{latentdiffusion,controlnet} and fidelity~\cite{dhariwal2021diffusion,imagen}.
Several studies have successfully applied diffusion models to synthesize training data for image classification \cite{syntheticimagenet}, object detection \cite{DiffusionEngine}, and fully-supervised segmentation \cite{wu2023datasetdm, freestyle}.
This raises the question: \textit{Can we also leverage the power of diffusion models to synthesize training data to further enhance the performance of scribble-supervised segmentation}?



In this work, we introduce ScribbleGen, a diffusion model conditioned on semantic scribbles to generate high-fidelity synthetic training images.
Deep image-generative models such as diffusion models commonly used today, often require large datasets to produce high-quality images.
This leads to a paradox where to upscale our training dataset, we need to already have access to a large training dataset.
We address this problem by including a new parameter in the generative process, the encode ratio, which trades off image diversity for image photorealism. 

Our contributions are summarized as follows:
\begin{itemize}
    \item We are the first to leverage denoising diffusion models for generative data augmentation for scribble-supervised semantic segmentation. Our approach produces a spectrum of synthetic images conditioned on scribbles using different guidance scales and encode ratios.
    \item We provide detailed analyses and propose several schemes to combine synthetic and real data effectively for scribble-supervised semantic segmentation. We also identify the limitations of naive data augmentation schemes that can harm segmentation performance relative to not using synthetic training data at all.
    \item We achieve state-of-the-art results in scribble-supervised semantic segmentation, closing the gap between weakly-supervised and fully-supervised models as shown in Fig. \ref{fig:ssl}. In particular, our framework significantly improves segmentation results in the low-data regime, where only a limited number of images are available.
\end{itemize}

\section{Related work}
\label{sec:relatedwork}

\paragraph{Synthetic training data}
Numerous efforts have been dedicated to leveraging synthetic data for training perception models. 
IT-GAN \cite{itgan} shows that GAN-generated samples can help classification models learn faster and improve performance. 
DatasetGAN \cite{Datasetgan}, BigDatasetGAN \cite{Li_2022_CVPR}, and {HandsOff \cite{yang2023ai}} employ GANs \cite{stylegan, biggan} for jointly generating synthetic images and their corresponding labels for segmentation tasks.

Recent advances in diffusion models have brought notable stability during training, robust synthesis capabilities \cite{dhariwal2021diffusion}, and enhanced controllability \cite{controlnet}. 
As a result, there has been a significant shift towards the use of diffusion models for data synthesis, including for image classification \cite{syntheticimagenet, kattakinda2022invariant}, object detection \cite{DiffusionEngine}, instance segmentation \cite{xie2023mosaicfusion}, and semantic segmentation \cite{nguyen2023dataset, wu2023datasetdm, li2023open, freemask}. 
For example, by fine-tuning an Imagen \cite{imagen} model on ImageNet \cite{imagenet}, \cite{syntheticimagenet} generates synthetic images from text prompts to improve the performance of image classification. 
Similarly, D3S \cite{kattakinda2022invariant} introduces a novel synthetic dataset specially designed to mitigate the foreground and background biases prevalent in real images. 
{\cite{nguyen2023dataset, wu2023datasetdm, wu2023diffumask} jointly generate synthetic images and associated mask annotation}, akin to DatasetGAN, using a StableDiffusion \cite{latentdiffusion} image-generative model. {GroundedDiffusion \cite{li2023open} further generates the triplet of image, mask, and texts to adapt the pretrained diffusion model for open-vocabulary segmentation.}
FreeMask \cite{freemask} utilizes FreestyleNet \cite{freestyle} to synthesize images conditioned on full mask annotations.

Our work diverges from these initiatives by focusing on sparse labels (e.g., scribbles) from real images as generative conditions, encouraging the creation of realistic and diverse synthetic images. 
While FreeMask \cite{freemask} similarly conditions synthetic images on real data annotations, our method uses sparse rather than dense annotations, allowing for broader applications where dense labeling is expensive.

\paragraph{Guidance in Diffusion models}

Diffusion models excel in various tasks due to their controllability~\cite{controlnet}. 
They're used to generate image content~\cite{ddpm}, image layout~\cite{latentdiffusion,selfguided,meng2021sdedit}, audio content~\cite{liu2023audioldm}, human motion~\cite{tevet2022human_mdm}, etc. 
Guidance signals can also be incorporated to enhance image fidelity~\cite{dhariwal2021diffusion, classifierfree} relative to unconditional generation.
It has been shown that diffusion models can be guided by pretraining a noisy-data-based classifier, known as Classifier-guidance~\cite{dhariwal2021diffusion}. 
On the other hand, classifier-free guidance~\cite{classifierfree} removes the need for extra pretraining by randomly dropping out the guidance signal during training. 
We develop a framework that utilizes classifier-free guidance for generative data augmentation to improve scribble-based segmentation.  

\paragraph{Weakly-supervised segmentation}
Weakly-supervised segmentation methods use weak annotations rather than full segmentation masks to train segmentation networks for images~\cite{rloss,ke2021universal,chang2020weakly,TEL,AGMM,Groupvit} or point clouds~\cite{scribblelidar}. 
Forms of weak annotations include points~\cite{whatsthepoint,rloss,AGMM}, scribbles~\cite{rloss,TEL,AGMM}, bounding boxes~\cite{simpledoesit}, image-level tags \cite{chang2020weakly}, and text~\cite{Groupvit}. 
These methods can be roughly categorized into two groups. 
The first group proposes various unsupervised or semi-supervised losses such as entropy loss~\cite{changmixup}, CRF loss~\cite{rloss}, and contrastive-learning losses~\cite{ke2021universal}. 
The second group iteratively refines full-mask pseudo-labels~\cite {simpledoesit,scribblesup} during training to mimic full supervision. 
Many weakly-supervised approaches rely on class activation maps (CAMs)~\cite{CAM,changmixup} that gives localization cues from classification networks. 
Our generative data augmentation approach complements any existing weakly-supervised segmentation methods, as we show improved performance of several methods with our synthetic data.

Weak annotations can also be provided as input for segmentation networks at test time for interactive segmentation~\cite{deepgrabcut,sam}. 
For example, Segment Anything~\cite{sam} allows prompts including clicks, bounding boxes, masks, or text. 
While Segment Anything~\cite{sam} provides many masks in a semi-automatic way for training interactive segmentation, we focus on synthetic image synthesis for training weakly-supervised segmentation.

\section{Method}

\begin{figure*}
    \centering
    \includegraphics[width=1.0\textwidth]{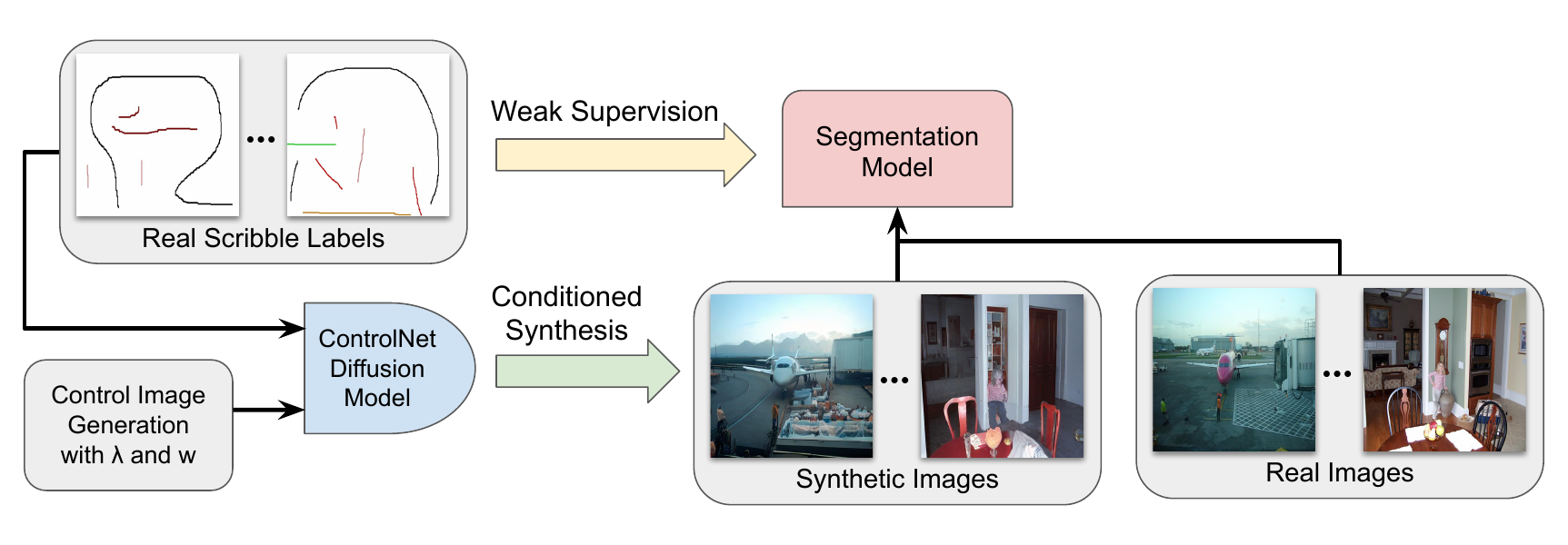}
    \vspace{-0.5cm}
    \caption{Given a limited number of real scribbles, we pretrain a ControlNet-based diffusion model for high-fidelity synthesis of images conditioned on scribbles. 
    We can control the image synthesis with the encode ratio $\lambda$ and the guidance scale $w$.
    These image-scribble pairs can then be smoothly integrated into the training of scribble-based semantic segmentation.}
    \label{fig:overall}
\end{figure*}

In this section, we describe our method of generative data augmentation for weakly-supervised semantic segmentation outlined in Fig. \ref{fig:overall}. 
First, in Sec. \ref{sec:background}, we provide a background on sampling from diffusion models.
Then, in Sec. \ref{sec:controlnet}, we introduce a variant of ControlNet~\cite{controlnet} conditioned on scribble labels and text prompts. 
We further discuss how to achieve semantically consistent images and trade off diversity and photorealism through guided diffusion and encode ratio in Sec. \ref{sec:guidance} and Sec. \ref{sec:encoderatio}, respectively. 
Sec. \ref{sec:data-aug} proposes several schemes to effectively combine synthetic and real images for training segmentation networks.

\subsection{Background}
\label{sec:background}

\paragraph{Diffusion models.} 
Diffusion models~\cite{sohl2015deep,ddpm,song2021scorebased_sde} learn to reverse a forward process that gradually adds noise to an image $\bxref$ until the original signal is fully diminished.
After training, following the reverse process allows us to sample an image $\bx_0$ given noise $\bx_T\sim\mathcal{N}(0, I)$.
Learning this reverse process reduces to learning a denoiser $\bepsilon_\theta$ that recovers the original image from a noisy image $\bx_t$ as 
\begin{equation}
\bxref\approx f_\theta(\bx_t,t):=(\bx_t - (1-\overline{\alpha}_{t})\bepsilon_{\btheta}(\bx_t, t))/\sqrt{\overline{\alpha}_{t}}.
\label{eq:reconstruct}
\end{equation}
To get high-quality samples, the standard diffusion model sampling process~\cite{ddpm} requires many (often $T=1,000$) neural function evaluations.
Using a non-Markovian forward process, Denoising Diffusion Implicit Model (DDIM)~\cite{ddim} samplers forego several intermediate function calls, accelerating sampling.
Let $\tau$ be an increasing subsequence of $[T,\ldots,1]$ and define the DDIM forward process for some stochasticity parameter $\sigma\in\mathbb{R}_{\ge0}^T$ as
\begin{multline} \label{eqn2:qsigma}
q_{\sigma}(\bx_{\tau_{i-1}}|\bx_{\tau_i},\bx_0)=
\mathcal{N}\Bigg(
\sqrt{\alpha_{\tau_{i-1}}}\bx_0\\
+\sqrt{1-\alpha_{\tau_{i-1}}-\sigma_{\tau_i}^2}\cdot\frac{\bx_{\tau_i}-\sqrt{\alpha_{\tau_i}}\bx_0}{\sqrt{1-\alpha_{\tau_i}}},\;\sigma_{\tau_i}^2I\Bigg).
\end{multline}
We can then sample from the generative process using the abovementioned forward process. 
In particular, using $f_\theta(x_t,t)$ as defined in Eq. (\ref{eq:reconstruct}) we can sample $\bx_{\tau_{i-1}}$ from $\bx_{\tau_i}$ by
\begin{equation}\label{eq:ddim-sampling}
p_\theta^{(\tau_i)}(\bx_{\tau_{i-1}}|\bx_{\tau_i})=
\begin{cases}
\mathcal{N}(f_\theta(\bx_{\tau_i}, \tau_i),~\sigma_{\tau_i}^2I)
& \text{if $i=1$}\\
q_{\sigma}(\bx_{\tau_{i-1}}|\bx_{\tau_i},f_\theta(\bx_{\tau_i},\tau_i))
& \text{if $i>1$}
\end{cases}
\end{equation}
We slightly abuse notation here and define $\tau_0=0$ so that when $i=1$, we sample the denoised image $\bx_0$.

\paragraph{Classifier-free guidance.}  
To trade off mode coverage and sample fidelity in a conditional diffusion model, \cite{dhariwal2021diffusion} proposes to guide the image generation process using the gradients of a classifier, with the additional cost of training the classifier on noisy images. 
To address this drawback, classifier-free guidance~\cite{classifierfree} does not require any classifier. They obtain a conditional and unconditional network combination in a single model by randomly dropping the guidance signal $\bc$ during training. 
After training, it empowers the model with progressive control over the degree of alignment between the guidance signal and the sample by varying the guidance scale $w$ when a larger $w$ leads to greater alignment with the guidance signal:
\begin{equation}\label{eq:cfg-default}
    \tilde{\bepsilon}_{\theta}(\bx_{t},t;~\bc, w) = (1+w) \bepsilon_{\theta}(\bx_{t},t;~\bc) - w \bepsilon_{\theta}(\bx_{t},t).
\end{equation}

\subsection{Scribble-conditioned Image Synthesis}
\label{sec:controlnet}

We consider a semantic synthesis approach to generating our synthetic training data.
The synthetic training data is generated conditioned on real segmentation labels from the training dataset. 
We leverage a typical denoising diffusion model, ControlNet~\cite{controlnet}, to achieve image synthesis conditioned on the segmentation scribbles.
Our model is trained using the usual DDPM~\cite{ddpm} object: given a noisy image $\bx_t$ (in reality $\bx_t$ is a latent representation as in~\cite{latentdiffusion}, but we omit this detail for brevity) and conditioning input $\bc$ it predicts the added noise $\bepsilon$.
Our segmentation scribbles on which the model is conditioned are represented as RGB images in $\mathbb{R}^{h\times w\times 3}$ with different colors for every class, though we explore other representations in Sec.~\ref{sec:exp-ablation}.

Finally, we note that it is difficult for the ControlNet model to produce semantically consistent images with the given scribble labels. 
We hypothesize that this is due to the difficulty of encoding class information in RGB images, especially in the early stages of training. 
Therefore, we supplement our model with text prompts that include all the classes within the image.
Adding these prompts significantly improves image class consistency and leads to higher-quality images relative to an unchanging default prompt.
We explore the effect of this prompt in Sec.~\ref{sec:exp-ablation}.

Our ControlNet training objective is thus
\begin{equation}
    \mathcal{L}_\mathrm{CN}(\btheta) = \mathbb{E}_{(\bxref, \bc_s, \bc_t), t, \bepsilon }\left[ \lVert \bepsilon - \bepsilon_\theta(\bx_t, t, \bc_s, \bc_t) \rVert _2^2 \right],
\end{equation}
where $(\bxref,\bc_s,\bc_t)$ is the triplet of the original (unnoised) image, the conditioning scribble label, and the conditioning text prompt and $\bepsilon_\theta$ is our ControlNet diffusion model.

\subsection{Classifier-free Scribble Guidance}
\label{sec:guidance}

We leverage diffusion guidance to further improve semantic consistency between the generated synthetic image and conditional input.
Following the proposals from Classifier-free Guided Diffusion~\cite{classifierfree}, we randomly drop out 10\% of all conditioning scribble inputs $\bc_s$, replacing them with a randomly initialized, learned embedding $\tilde{\bc}$, when training the ControlNet model. 
By modifying Eq. \ref{eq:cfg-default}, we arrive at a new guided noise prediction function:
\begin{equation}\label{eq:cfg-custom}
\tilde{\bepsilon}_{\theta}(\bx_{t},t;\bc_s,\bc_t, w) = (1+w) \bepsilon_{\theta}(\bx_{t},t;\bc_s,\bc_t) - w \bepsilon_{\theta}(\bx_{t},t;\tilde{\bc}).
\end{equation}
While ControlNet uses a pre-trained Stable-Diffusion model~\cite{latentdiffusion}, which is trained conditionally and unconditionally, scribble drop-out during training can be viewed as finetuning the unconditional diffusion model to our dataset.
We have found that the guidance scale, $w$, can significantly impact the quality of generated images, especially with respect to the fine-grain details of the produced image. 
We further ablate this hyperparameter's impact in Sec \ref{sec:exp-ablation}.

\subsection{Control Image Diversity via Encode Ratio}
\label{sec:encoderatio}

\begin{figure*}
    \centering
    \begin{subfigure}[t]{0.7\textwidth}
        \centering
        \includegraphics[width=1.0\textwidth,valign=c]{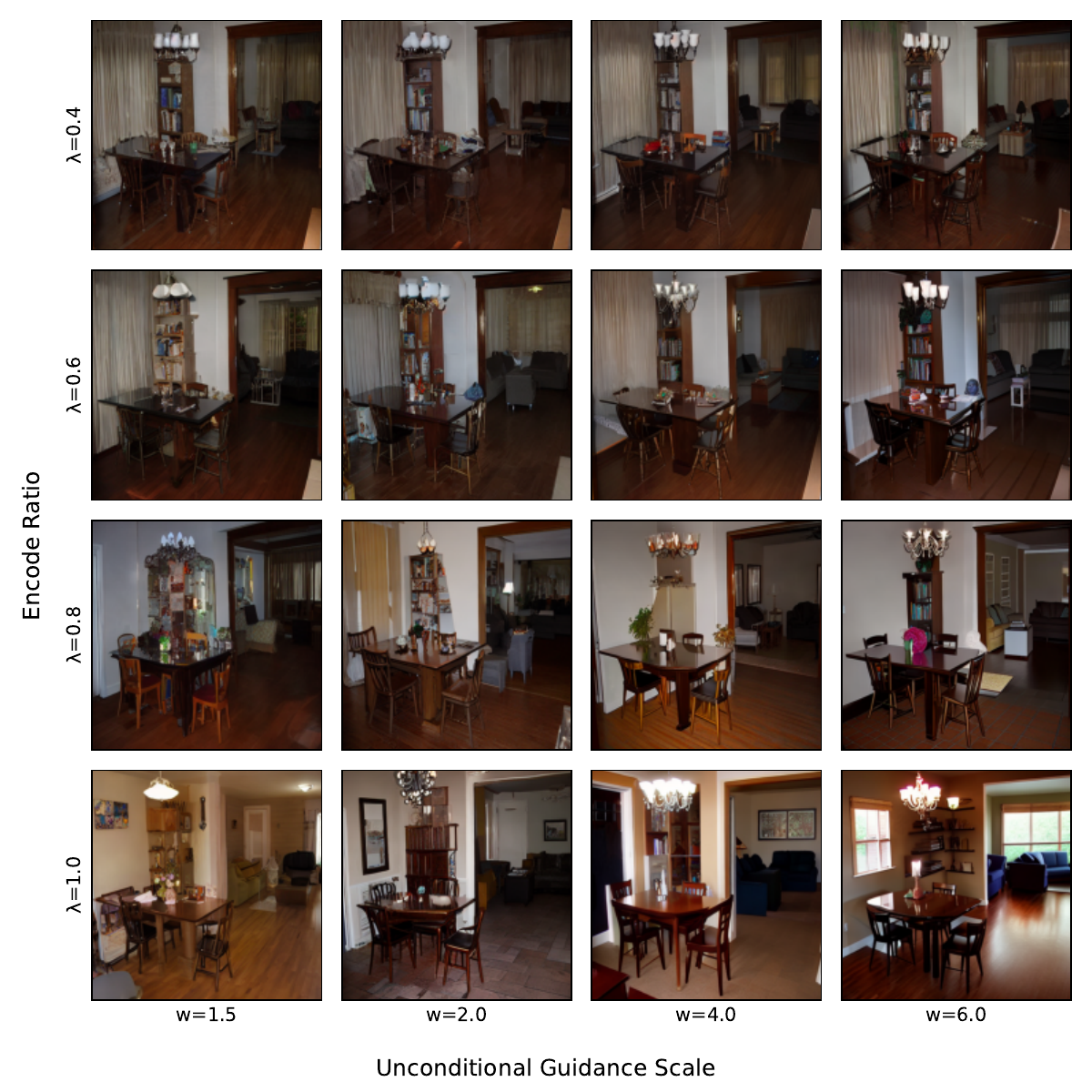}
    \end{subfigure}
    \begin{subfigure}[t]{0.25\textwidth}
        \centering
        \includegraphics[width=1.0\textwidth,valign=c]{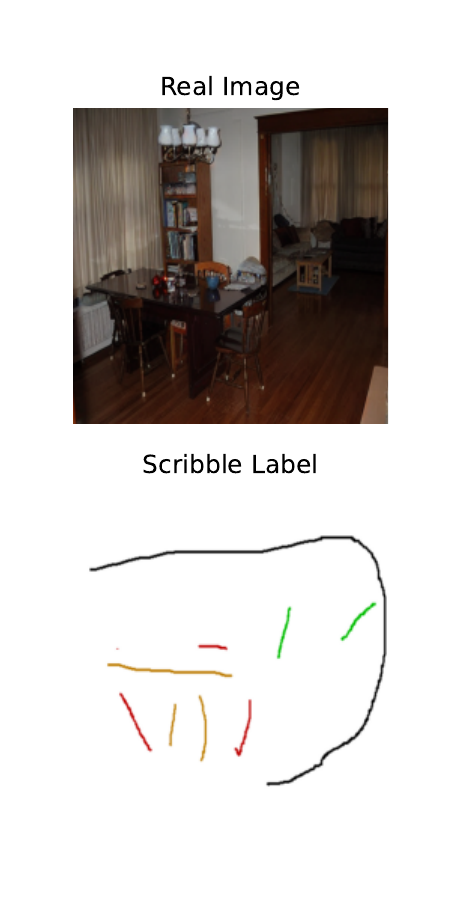}
    \end{subfigure}
    \vspace{-0.3cm}
    \caption{
    Left: Our sampled synthetic images conditioned on the ground-truth scribble. By sampling using different guidance scales and encode ratios we are able to generate a whole spectrum of realistic synthetic training images.
    Right: The ground-truth real image and corresponding scribble label.
    }
    \label{fig:sampling}
\end{figure*}

The vanilla diffusion model denoises sampled Gaussian noise $\bx_T\sim \mathcal{N}(0,I)$ iteratively until $\bx_0$ at inference time.
In practice, synthetic images generated this way may be unrealistic, particularly when training data is limited for our scribble-conditioned diffusion model. 
To improve photorealism at the cost of diversity, we propose another forward diffusion process parameter, the encode ratio $\lambda\in(0,1]$. 
Specifically, we perform $\lambda\cdot T$ noise-adding forward diffusion steps to the input images and, during inference, denoise $\bx_{\lambda T}$ iteratively until $\bx_0$.
Thus, for $\lambda=1$, there is no change, but for small choices of $\lambda$, there is less noise added to the image $\bx_0$.
As $\lambda\to 0$, the sampled image will become increasingly similar to the original $\bxref$.
Therefore, a whole spectrum of synthetic images with varying levels of similarity to the reference image can be achieved by varying our choice of $\lambda$.
We outline our sampling algorithm, which combines the accelerated DDIM sampling from Sec. \ref{sec:background}, the scribble guidance from Sec. \ref{sec:guidance}, and the encode ratio from Sec \ref{sec:encoderatio} in Algorithm \ref{alg:sampling}.
Fig. \ref{fig:sampling} shows synthetic images generated with varying guidance scales and encode ratios.

\begin{algorithm}
\caption{Conditional DDIM sampling with guidance scale $w$ and encode ratio $\lambda$}
\label{alg:sampling}
\DontPrintSemicolon
\Require{$q_\sigma$: forward process}
\Require{$\bxref$: a reference image}
\Require{$w\ge0$: guidance scale}
\Require{$\lambda\in [0, 1]$: encode ratio}
\Require{$N\in\{1,\ldots,T\}$: number of reverse diffusion process steps}
\Require{$c_t$ and $c_s$: text prompt and scribble conditioning}
$\bepsilon\sim\mathcal{N}(0,I)$\;
$\tau=[\lfloor\frac{\lambda T}{N}n\rfloor:0\le n\le N]$\;
\Comment{Note if $\lambda=1$ then $\bx_{\tau_N}\sim\mathcal{N}(0,I)$}
$\bx_{\tau_N}=\sqrt{\alpha_{\tau_N}}\bxref+\sqrt{1-\alpha_{\tau_N}}\bepsilon$\;
\For{$i=N$ to $1$}{
    \Comment{Predict added noise using diffusion guidance (\ref{eq:cfg-custom})}
    \mbox{$\tilde{\bepsilon}_{\tau_{i}}=(1+w)\bepsilon_\theta^{(\tau_{i})}(\bx_{\tau_{i}},c_t,c_s)-w\bepsilon_\theta^{(\tau_{i})}(\bx_{\tau_i}, c_t,\tilde{c}_s)$}\;\vspace{-.9em}
    \Comment{\mbox{Accelerated DDIM sampling (\ref{eq:ddim-sampling})}}
    $\hat{\bx}_0=(\bx_{\tau_i}-\sqrt{1-\alpha_{\tau_i}}\cdot\tilde{\bepsilon}_{\tau_i})/\sqrt{\alpha_{\tau_i}}$\;
    \uIf{$i=1$}{
        $\bx_{0}\sim \mathcal{N}(\hat{\bx}_{0},\sigma^2_{\tau_1}I)$\;
    }
    \Else{
        $\bx_{\tau_{i-1}}\sim q_\sigma(\bx_{\tau_{i-1}}|\bx_{\tau_i},\hat{\bx}_0)$\;
    }
}
\Return $\bx_0$\;
\end{algorithm}

\subsection{Combine synthetic images with real images}
\label{sec:data-aug}

Generative data augmentation can, in principle, produce an infinite amount of synthetic images. 
However, naively combining real and synthetic images can harm rather than benefit weakly-supervised segmentation models, as we have observed. 
In particular, it is not clear which choices of the guidance scale $w$ and encode ratio $\lambda$ are optimal. 
We choose the optimal guidance scale, as determined in Sec.~\ref{sec:exp-ablation}. 
For encode ratio $\lambda$, we propose and systematically evaluate two strategies for combining synthetic with real images.

Let $\mathcal{X}=\{\mathbf{x}_1,\ldots,\mathbf{x}_n\}$ denote the set of all real images and $\mathcal{Y}=\{\mathbf{y}_1,\ldots,\mathbf{y}_n\}$ denote the set of all (scribble) labels. 
Then we produce a set of synthetic images $\hat{\mathcal{X}}=\{\hat{\mathbf{x}}_1,\ldots,\hat{\mathbf{x}}_n\}$ where $\hat{\mathbf{x}}_i=\mathrm{DM}_\theta(\mathbf{y}_i,\bc_i;w,\lambda)$ is the output of our trained diffusion model, $\mathrm{DM}_\theta$, conditioned on the scribble $\mathbf{y}_i$ and prompt-condition $\bc_i$, given guidance scale $w$ and encode ratio $\lambda$.
We may then produce a new, augmented dataset $\mathcal{X}'=\mathrm{concat}(\mathcal{X},\hat{\mathcal{X}})$ and $\mathcal{Y}'=\mathrm{concat}(\mathcal{Y},\mathcal{Y})$. 
Note this means each label, $\mathbf{y}_i$, appears twice in our dataset, once for the real image $\mathbf{x}_i$ and once for the synthetic image $\hat{\mathbf{x}}_i$. 

\begin{itemize}
    \item \textbf{Fixed encode ratio $\lambda$}: We choose a fixed encode ratio which gives a fixed synthetic dataset $\hat{\mathcal{X}}$. Using the default value of $\lambda=1$ yields the most diverse synthetic images with possibly inferior image fidelity. We find the optimal $\lambda$ that gives the best segmentation in our experiments.
    \item \textbf{Adaptive encode ratio $\lambda$}: To avoid hyper-parameter search, we also propose an adaptive scheme for choosing $\lambda$. We gradually increase the encode ratio $\lambda$ while training downstream segmentation networks, similar to curriculum learning. Initially, synthetic images used for training are similar to real images, which are considered an easier curriculum to learn. Synthetic images diverge increasingly from the real images as training progresses.
    For this case, the synthetic dataset is formed at epoch $e$ as $\hat{\mathcal{X}}=\{\hat{\bx}_{1,\lambda_e},\ldots,\hat{\bx}_{1,\lambda_e}\}$ where we follow the encode ratio schedule $[\lambda_1,\ldots,\lambda_E]\in\Lambda^E$ where $E$ is the number of training epochs.
\end{itemize}

\vspace{-0.25cm}

\section{Experiments}
\label{sec:exp-details}


Sec.~\ref{sec:exp-mainresults} summarize our main results that show improvements on several scribble-supervised segmentation methods using our generative data augmentation. 
In Sec.~\ref{sec:exp-ssl}, we further explore the challenging scenario with limited number of real images. 
We show that naive implementations of generative data augmentation may harm the performance, whereas our data augmentation scheme improves. 
Sec. \ref{sec:exp-ablation} gives an ablation study on guidance scale and encode ratio, two critical degrees of freedom for our image synthesis.

\paragraph{Dataset and Implementation Details} We report results on the standard PASCAL VOC12 segmentation dataset which contains 10~582 images for training and 1~449 images for validation. We utilize scribbles from ScribbleSup dataset~\cite{scribblesup} with only 3\% pixels labeled on average.

For image synthesis, we use a latent diffusion model~\cite{latentdiffusion} with a downsampling rate of $f=8$, so that an input image of size $512\times512$ is downsampled to $64\times64$. We use Stable Diffusion 1.5 as the backbone for ControlNet~\cite{controlnet} and finetune ControlNet for 200 epochs with a batch size of 16 using two A100 80GB GPUs. 
We set $T=1000$ discrete timesteps for ControlNet and use a linear learning rate scheduler from an initial rate of $10^{-4}$ during training. 
For scribble conditioning, we randomly dropout 10\% of scribbles, replacing them with a learned embedding of the same size. 
Scribble labels are represented as RBG images in $\{1,\ldots,255\}^{512\times512\times3}$. 
We also provide the text prompt "a high-quality, detailed, and professional image of [list of classes]" as suggested in \cite{controlnet}.
We provide visualizations of our synthetic dataset in the supplementary material.


\hspace{-0.15in}\textbf{Evaluation metric.}
We evaluate both the diversity and fidelity of the generated images by the Fréchet Inception Distance (FID) \cite{heusel2017gans_fid}, as it is the \textit{de facto} metric for the evaluation of generative methods, e.g., \cite{dhariwal2021diffusion,stylegan,biggan,imagen}. It provides a symmetric measure of the distance between two distributions in the feature space of Inception-V3~\cite{szegedy2016rethinking_inceptionv3}. 
We use FID as our primary metric for the sampling quality. 
We realize, however, that FID should not be the only metric for evaluating the downstream impact of synthetic data for training segmentation networks. 
Hence, we also report segmentation results trained with synthetic data only to evaluate synthetic data, similar to the Classification Accuracy Score (CAS) proposed by \cite{classaccscore} but for semantic segmentation.
We report the standard mean Intersection Over Union (mIOU) metric for segmentation results.

\subsection{Generative data augmentation improves scribble-supervised semantic segmentation}
\label{sec:exp-mainresults}

For our experiments, we consider two methods of weakly-supervised semantic segmentation, including simple regularized losses (RLoss)~\cite{rloss} and the current state-of-the-art in scribble-supervised segmentation, Adaptive Gaussian Mixture Models (AGMM)~\cite{AGMM}. For both methods, we jointly train them on the original training set and our augmented training set.
Both methods also follow a polynomial learning rate scheduler.
The sampling of synthetic training images is outlined in Sec.~\ref{sec:data-aug}. Table~\ref{tab:main-results} shows improved results using generative data augmentation for both RLoss and AGMM. Our method with synthetic data further reduces the gap between weakly-supervised and fully-supervised segmentation.
We show visualizations of our segmentation results with and without using our generative data augmentation in Fig.~\ref{fig:segmentation}.
We also include further visualizations in the supplementary material.

\begin{table*}[h]
    \centering
    \begin{tabular}{lcccll}
    \toprule
        Method & Network & Supervision & Synthetic Data & Augmentation Scheme & mIoU (\%) \\
        \midrule
        (1) *DeeplabV3+ \cite{chen2018deeplab} & MobileNet \cite{mobilenet} & Full mask &  & -- & 72.1 \\
        (2) *DeeplabV3+ \cite{chen2018deeplab} & ResNet101 \cite{resnet} & Full mask &  & -- & 79.3 \\
        \midrule
        RLoss \cite{rloss} & (1) & Scribble &  & -- & 68.4 \\
        RLoss \cite{rloss} & (1) & Scribble & \checkmark & Fixed $\lambda=1.0$ & 69.4 (\green{+1.0}) \\
        RLoss \cite{rloss} & (1) & Scribble & \checkmark & Fixed $\lambda=0.5$ & \textbf{70.0} (\green{+1.6}) \\
        RLoss \cite{rloss} & (2) & Scribble &  & -- & 76.6 \\
        RLoss \cite{rloss} & (2) & Scribble & \checkmark & Fixed $\lambda=1.0$ & {76.1} (\red{-0.5}) \\
        RLoss \cite{rloss} & (2) & Scribble & \checkmark & Fixed $\lambda=0.7$ & \textbf{77.0} (+\green{0.4}) \\
        \midrule
        AGMM \cite{AGMM} & (2) & Scribble &  & -- & 76.4 \\
        *AGMM \cite{AGMM} & (2) & Scribble &  & -- & 78.1 \\
        AGMM \cite{AGMM} & (2) & Scribble & \checkmark & Fixed $\lambda=1.0$ & {78.0} (\red{-0.1}) \\
        AGMM \cite{AGMM} & (2) & Scribble & \checkmark & Adaptive $\lambda$ & 78.7 (\green{+0.6}) \\
        AGMM \cite{AGMM} & (2) & Scribble & \checkmark & Fixed $\lambda=0.4$ & \textbf{78.9} (\green{+0.8}) \\
     \bottomrule
    \end{tabular}
    \caption{Generative data augmentation improves scribble-supervised semantic segmentation methods including RLoss~\cite{rloss} and AGMM~\cite{AGMM} on PascalVOC \cite{pascalvoc}. The best results are shown in \textbf{bold}. Numbers in parenthesis are relative improvement / decrease in comparison to the baseline without synthetic data. Note that * AGMM is our re-implementation which gives better results than reported~\cite{AGMM}.}
    \label{tab:main-results}
\end{table*}

\subsection{Low-data Regime Results}
\label{sec:exp-ssl}

For the low-data regime, we only consider the RLoss method due to its simplicity and speed to train.
We consider three different reduced datasets with 50\%, 25\%, and 12.5\% of all training images used, respectively.
For each of these cases, we train a ControlNet diffusion model on the limited dataset (following the same experimental setup described at the start of Sec.~\ref{sec:exp-details}) and sample synthetic images as usual.
The results of training RLoss on each of the reduced datasets for our different proposed augmentation schemes are reported in Fig.~\ref{fig:ssl}. 

\begin{figure}
    \centering
    \includegraphics[width=0.5\textwidth]{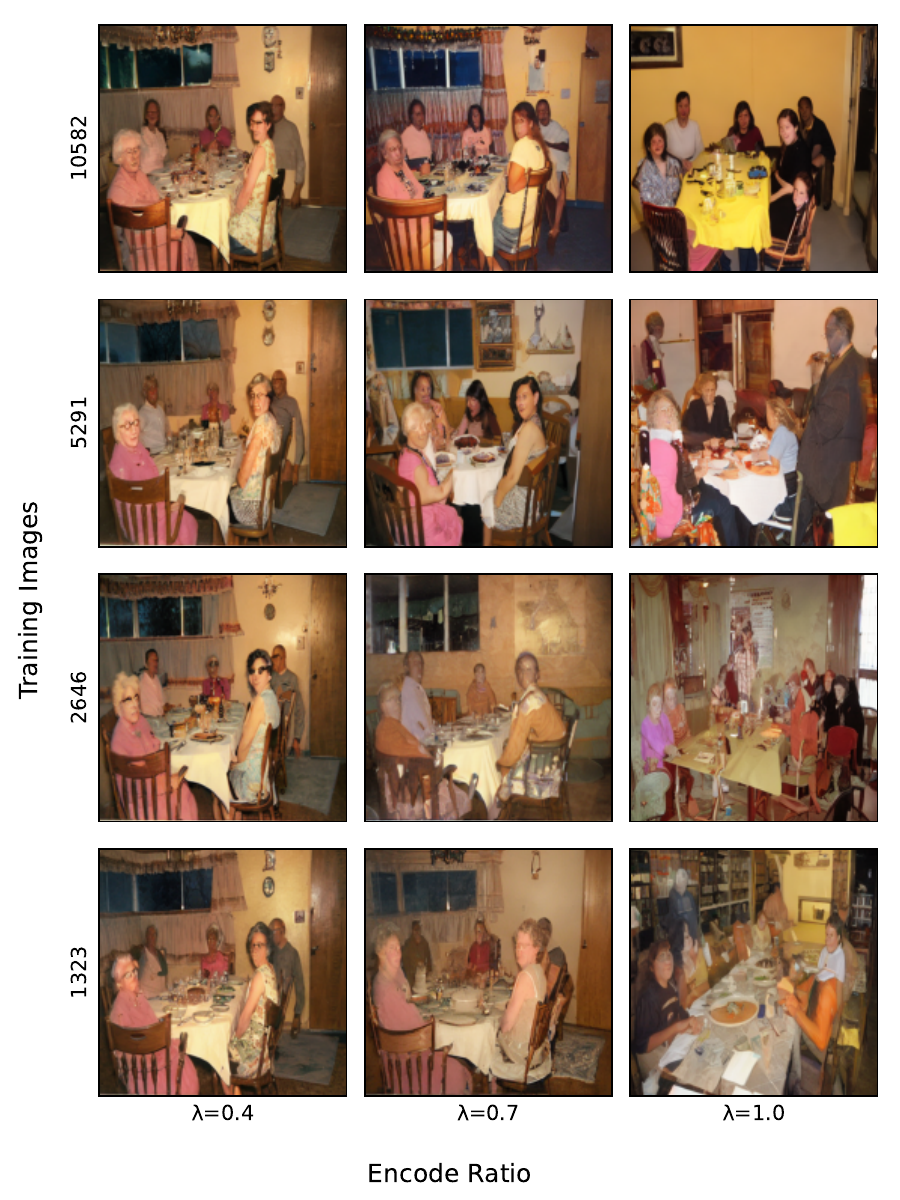}
    \vspace{-0.7cm}
    \caption{Synthetic images sampled from diffusion models with different numbers of training images and encode ratios $\lambda$.}
    \label{fig:sampling-ssl}
    \vspace{-0.2cm}
\end{figure}

We notice that the naive data augmentation fails to help in all of our reduced datasets and instead reduces model performance in all but the 50\% case.
Conversely, our proposed \textit{Adaptive $\lambda$ Sampling} improves or matches performance for all four datasets. 
We hypothesize this is due to the lack of training images required to ensure high-quality generation from our diffusion model.
This hypothesis is confirmed by the significantly higher FID scores for synthetic datasets generated with limited training data reported in Fig.~\ref{fig:ablation} middle.
We also confirm this hypothesis qualitatively in Fig.~\ref{fig:sampling-ssl}, where we observe that fully synthetic images deteriorate in quality as the number of training images decreases.
However, we can stabilize this deterioration by decreasing the encode ratio $\lambda$ to improve image realism.
Using our \textit{Adaptive $\lambda$ sampling}, the most synthetic (and thus lowest quality) images cannot impact model training as significantly due to the reduced learning from our scheduler.

\subsection{Ablation Studies}
\label{sec:exp-ablation}

\begin{figure*}
    \begin{subfigure}{.33\textwidth}
        \centering
        \includegraphics[width=0.8\textwidth]{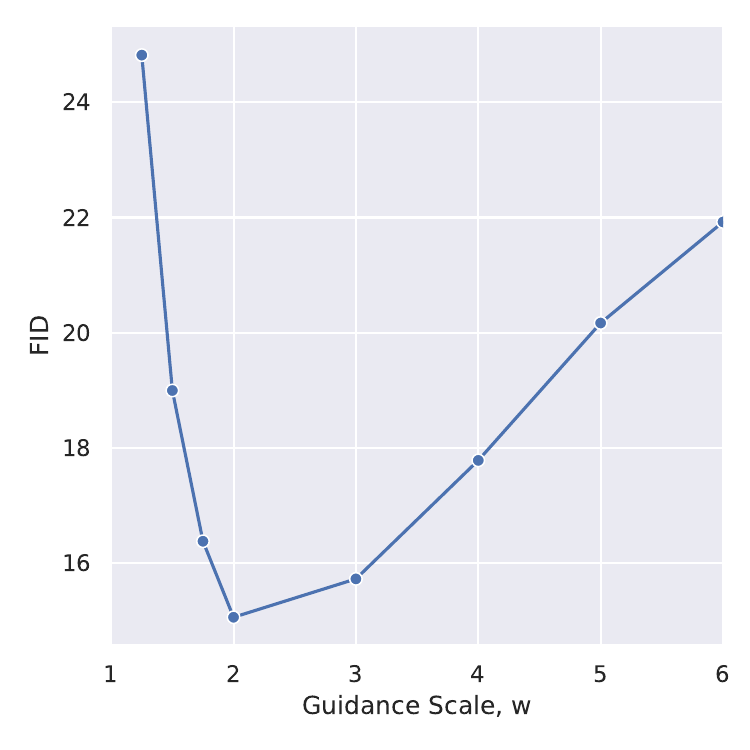}
    \end{subfigure}
    \begin{subfigure}{.33\textwidth}
        \centering
        \includegraphics[width=0.8\textwidth]{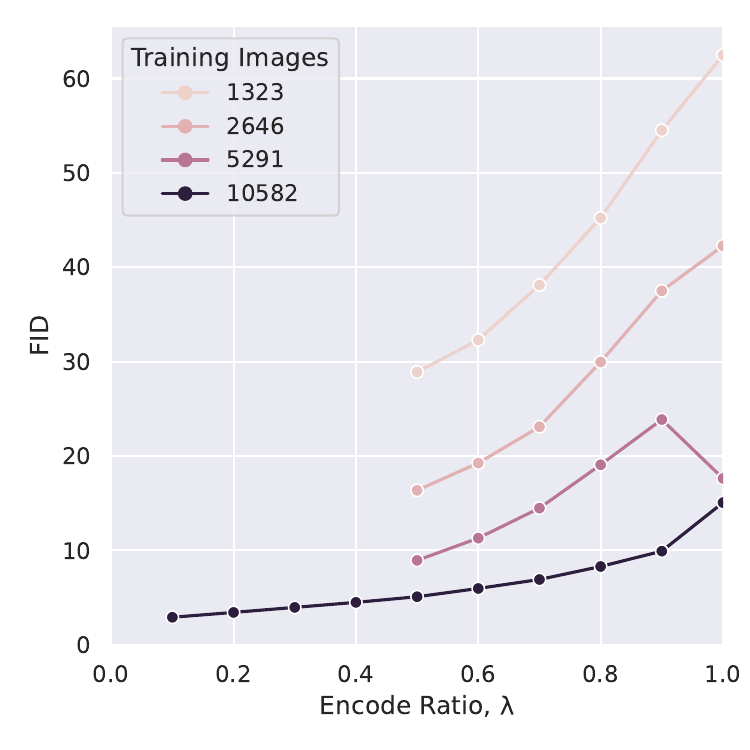}
    \end{subfigure}
    \begin{subfigure}{.33\textwidth}
        \centering
        \includegraphics[width=0.8\textwidth]{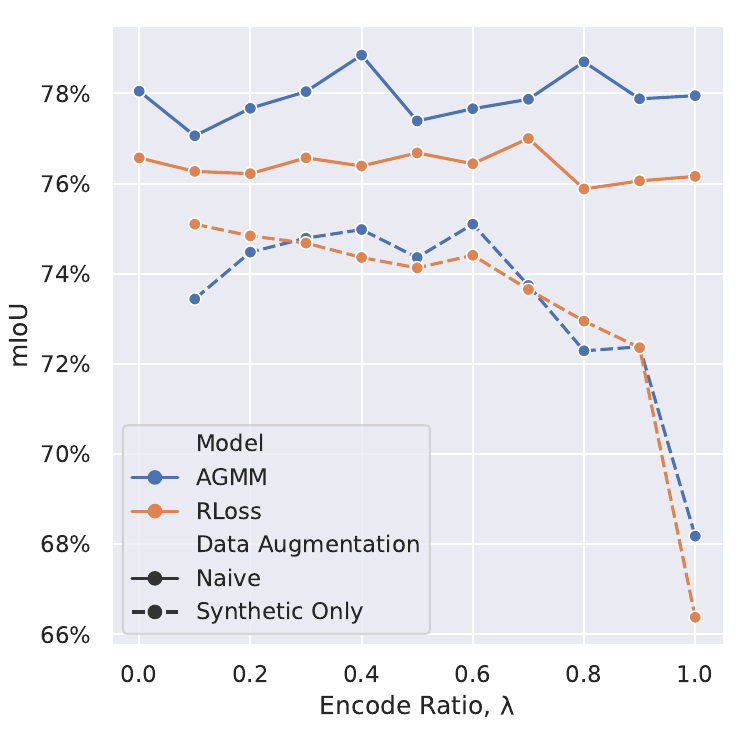}
    \end{subfigure}
    \caption{
    Left: The FID of our full training dataset when generated with different classifier-free guidance scales. 
    Results are reported for ControlNet trained all 10582 images.
    Middle: The FID of our training dataset when generated with different encode ratios. 
    Results are reported for four ControlNet models trained on a different number of images.
    Right: The mIoU of a downstream segmentation model when trained on datasets of varying encode ratios.
    Note $\lambda=0.0$ corresponds to training on real images only.
    Results are reported for training on both naive data augmentation and only on synthetic images.
    In both cases, we use all 10582 images for training.
    }
    \label{fig:ablation}
\end{figure*}

\begin{figure*}[h]
\newcommand{\makedemoline}[1]{\includegraphics[width=0.15\linewidth]{images/demos/#1_img.jpg} & \includegraphics[width=0.15\linewidth]{images/demos/#1_baseline.png} & \includegraphics[width=0.15\linewidth]{images/demos/#1_full_resize.png} & 
\includegraphics[width=0.15\linewidth]{images/demos/#1_mixsig04.png} & 
\includegraphics[width=0.15\linewidth]{images/demos/#1_gt.png}
}
    \centering
    \begin{tabular}{ccccc}
        image &  AGMM~\cite{AGMM} & \footnotesize{Fully-supervised} 
        & \footnotesize{AGMM (+syn. images)} & Ground truth \\
        \makedemoline{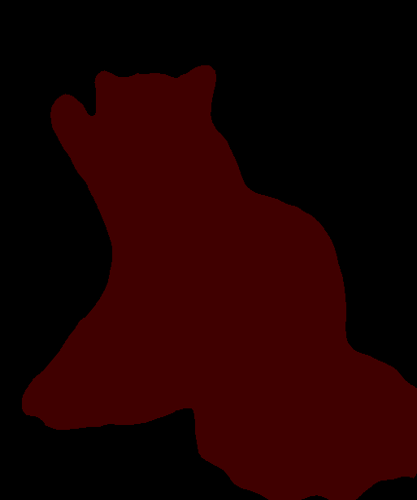} \\
        \makedemoline{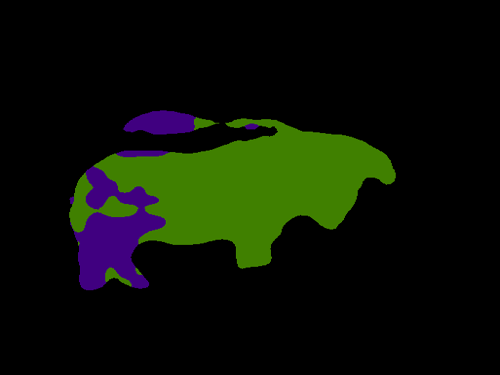} \\ 
    \end{tabular}
    \vspace{-2mm}
    \caption{Qualitative results on PASCAL dataset. Our generative data augmentation method improves scribble-supervised semantic segmentation methods such as AGMM~\cite{AGMM}.}
    \label{fig:segmentation}
\end{figure*}

\paragraph{Guidance Scale} We report the FID scores of our fully synthetic ($\lambda=1$) datasets as generated by our model trained on all PascalVOC training images in Fig.~\ref{fig:ablation} left.
This ablation study is how we decided to use $w=2$ for all other experiments, as it yields optimal FID.
We include further visualizations of the impact of the guidance scale on image synthesis in our supplementary material.

\paragraph{Encode Ratio} We report the FID scores of our diffusion models trained with a variable number of images as a function of the encode ratio $\lambda$ in Fig.~\ref{fig:ablation} middle.
We observe that the FID increases significantly as the number of training images decreases.
However, we can reduce the effect of limited training data by decreasing the encode ratio to promote image realism. 
This effect is most pronounced for the 1323 image-trained diffusion model, where we reduce the FID score by over 30 points by lowering the encode ratio.

We also evaluate segmentation model performance on synthetic data of varying encode ratios and report the final mIoU in Fig.~\ref{fig:ablation} right.
For these experiments, we train segmentation models training using the Fixed $\lambda$ data augmentation proposed in Sec.~\ref{sec:data-aug} and training exclusively on synthetic training data (i.e., $\mathcal{X}'=\hat{\mathcal{X}}$), akin to CAS \cite{classaccscore}.
We observe that the impact of varying the encode ratio $\lambda$ is limited in the data augmentation case but much more significant for the synthetic-only case.
We suppose that for the synthetic-only case, the quality of the synthetic images is more important, so decreasing the encode ratio to improve data realism matters more than data diversity.
We include further visualizations of the impact of the encode ratio on image synthesis in our supplementary material.

\paragraph{Conditioning Input} We also ablate modifying the conditioning input to ControlNet.
We try representing scribble labels as one-hot embeddings in $\{0,1\}^{h\times w\times C}$ where there are $C$ total classes.
Using these one-hot embeddings, we obtained a higher FID by 4.4 points relative to RGB embeddings, but we found no improvement in mIoU results using our Fixed $\lambda$ augmentation scheme. 
We also try using text prompts that don't include the classes in the image.
Using unchanging prompts (i.e., ``a high-quality, detailed, and professional image'') yields lower FID by 3.1 points relative to prompts that include the classes in the image and 1.9\% lower mIoU using our Fixed $\lambda$ augmentation scheme. 



\section{Conclusion and Future Work}

We propose leveraging diffusion models conditioned on scribbles to produce high-quality synthetic training data for scribble-supervised semantic segmentation. We advocate the use of classifier-free guided diffusion and introduce the encode ratio to control the generative process, allowing us to generate a spectrum of images.
We report state-of-the-art performance on scribble-supervised semantic segmentation with our generative data augmentation.

In the future, it will be interesting to train generative models for open-vocabulary image synthesis conditioned on sparse annotations. Our generative data augmentation has the potential to improve semi-supervised segmentation. We are also interested in end-to-end training of generative data augmentation and perception models, as metrics like FID are loosely related to perception performances.

\paragraph{Acknowledgement} The authors would like to thank Prof. Yuri Boykov, Prof. Olga Veksler, and Prof. Ming-Hsuan Yang for their helpful discussion and comments that improved the quality of this work.


{
    \small
    \bibliographystyle{ieeenat_fullname}
    \bibliography{main}
}

 \clearpage
\setcounter{page}{1}
\maketitlesupplementary

\section{Validation Set FID Results}
\label{sec:val-fid}

\begin{figure}
    \centering
    \includegraphics[width=0.45\textwidth]{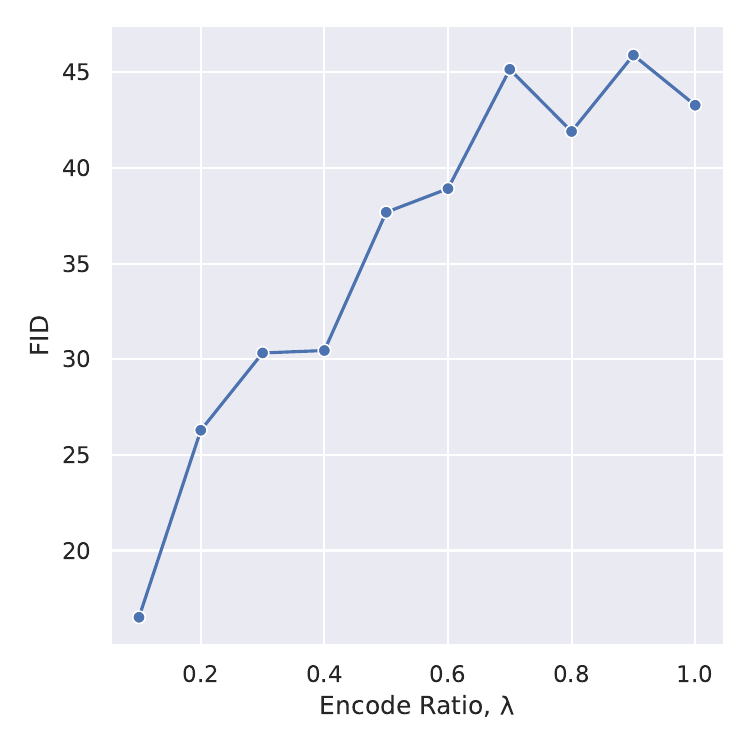}
    \caption{FID reported on the validation set. Synthetic images are from our ControlNet model trained on all of PascalVOC. Images are synthesized conditioned on scribbles from the validation set with varying encode ratios.}
    \label{fig:val-fid}
\end{figure}

In this section, we report the FID of our synthetic images on the validation set.
To achieve this, we provide scribbles from the validation set of PascalVOC as conditioning input to our trained ControlNet models.
Since the ControlNet models were not trained with data from the validation set, these are previously unseen scribbles.
In Table~\ref{tab:val-fid}, we report the impact of the number of training images of the ControlNet model on validation FID.
In Fig.~\ref{fig:val-fid}, we report the impact of the encode ratio on validation FID.

\begin{table}[]
    \centering
    \begin{tabular}{cc}
    \toprule
    Training Images & FID \\
    \midrule
    10582 & 43.3\\
    5291 & 53.7\\
    2646 & 57.1\\
    1323 & 58.0\\
    \bottomrule
    \end{tabular}
    \caption{FID reported on the validation set. Synthetic images are from our ControlNet model with a varying number of training images. Synthetic images are conditioned on scribbles from the validation set, previously unseen to our model.}
    \label{tab:val-fid}
\end{table}

\section{Additional Qualitative Results}
\label{sec:qualitative}

\begin{figure*}
    \centering
    \includegraphics[width=0.9\textwidth]{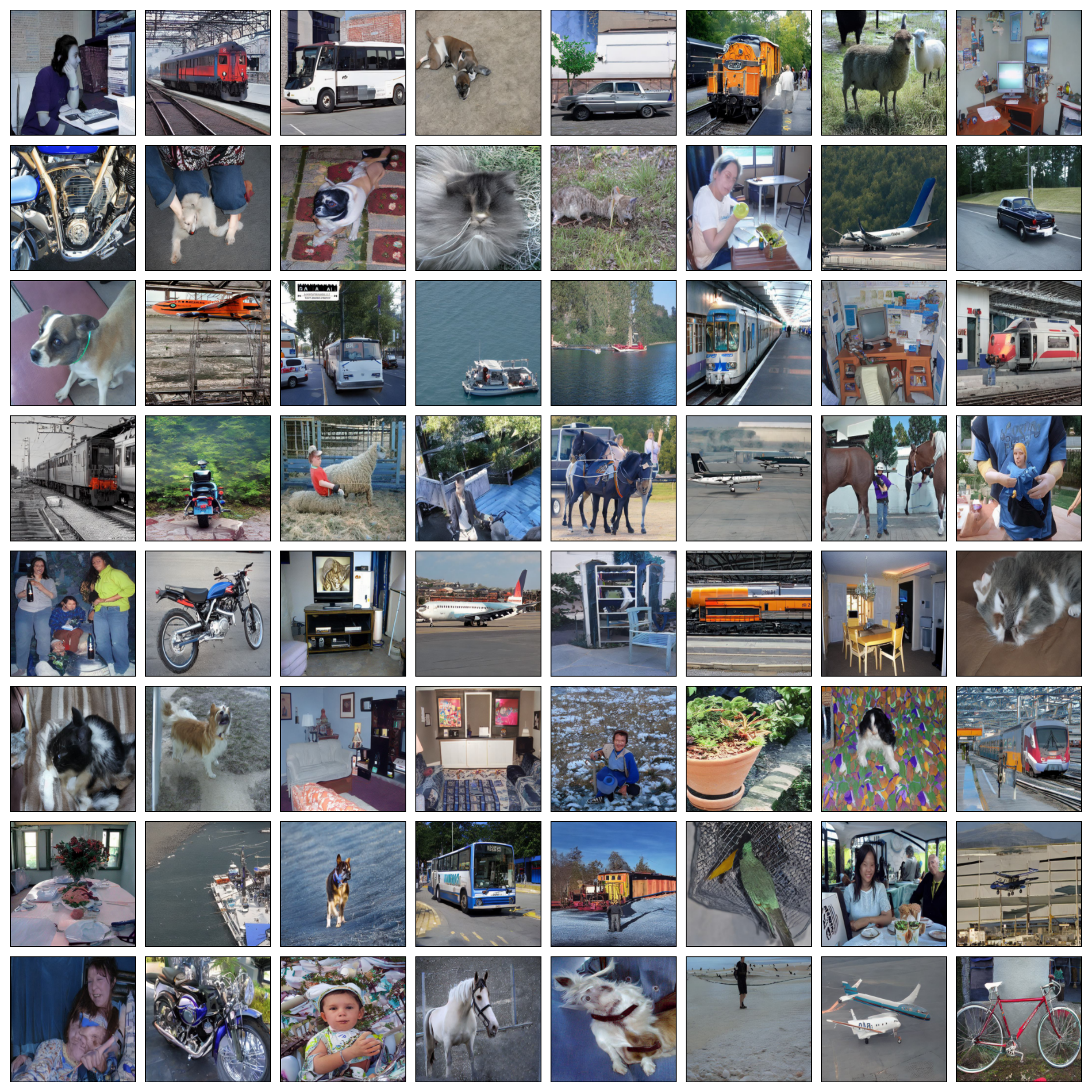}
    \caption{Synthetic training images sampled from a ControlNet model trained on all of scribble-supervised PascalVOC. All images are sampled using guidance scaled $w=2.0$ and encode ratio $\lambda=1.0$.}
    \label{fig:general-sampling}
\end{figure*}

\begin{figure*}
    \centering
    \includegraphics[width=0.9\textwidth]{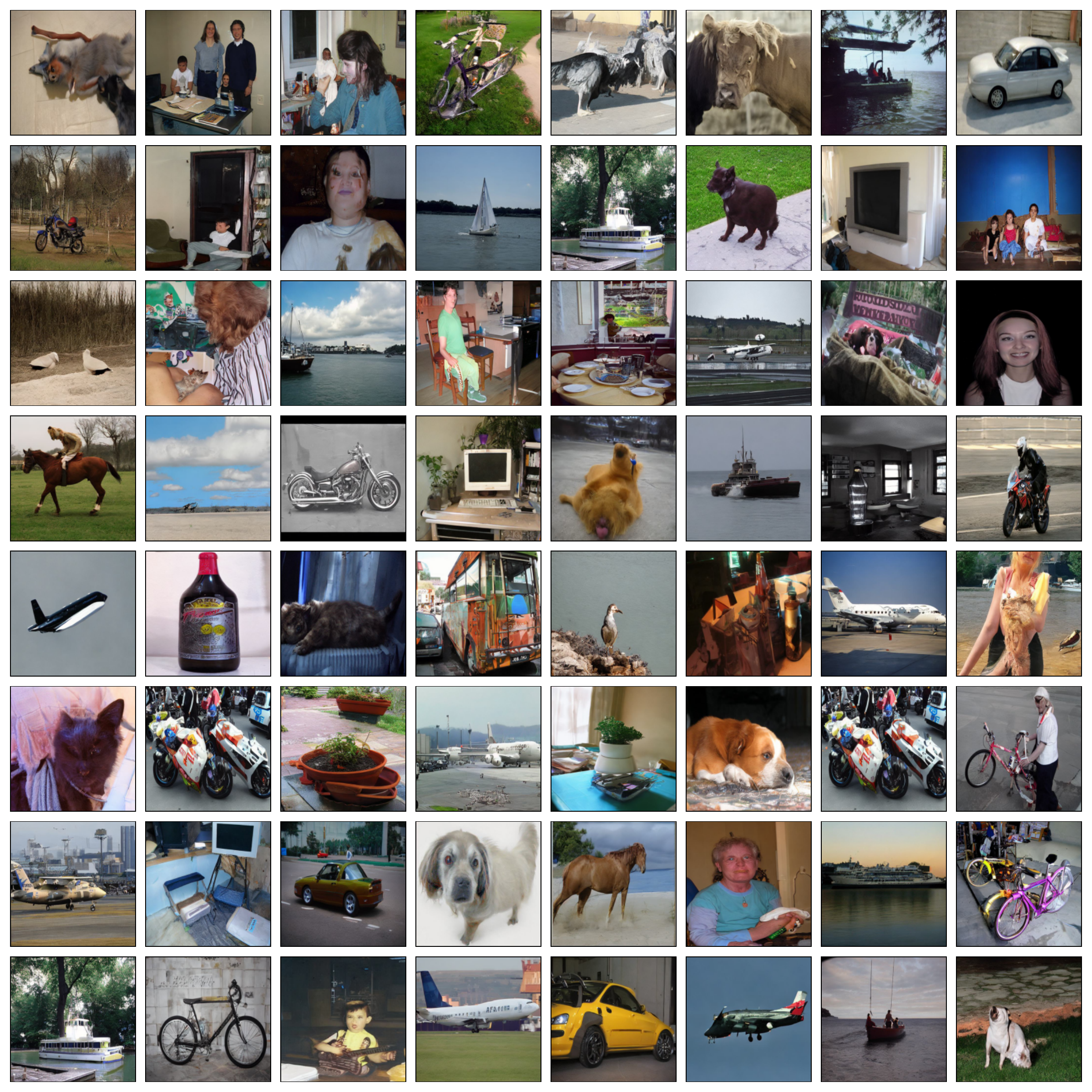}
    \caption{Synthetic validation images sampled from a ControlNet model trained on all of scribble-supervised PascalVOC. Images are synthesized conditioned on scribbles from the validation set, which the ControlNet model has not been trained on. All images are sampled using guidance scaled $w=2.0$ and encode ratio $\lambda=1.0$.}
    \label{fig:validation-sampling}
\end{figure*}

\begin{figure*}
    \centering
    \includegraphics[width=0.9\textwidth]{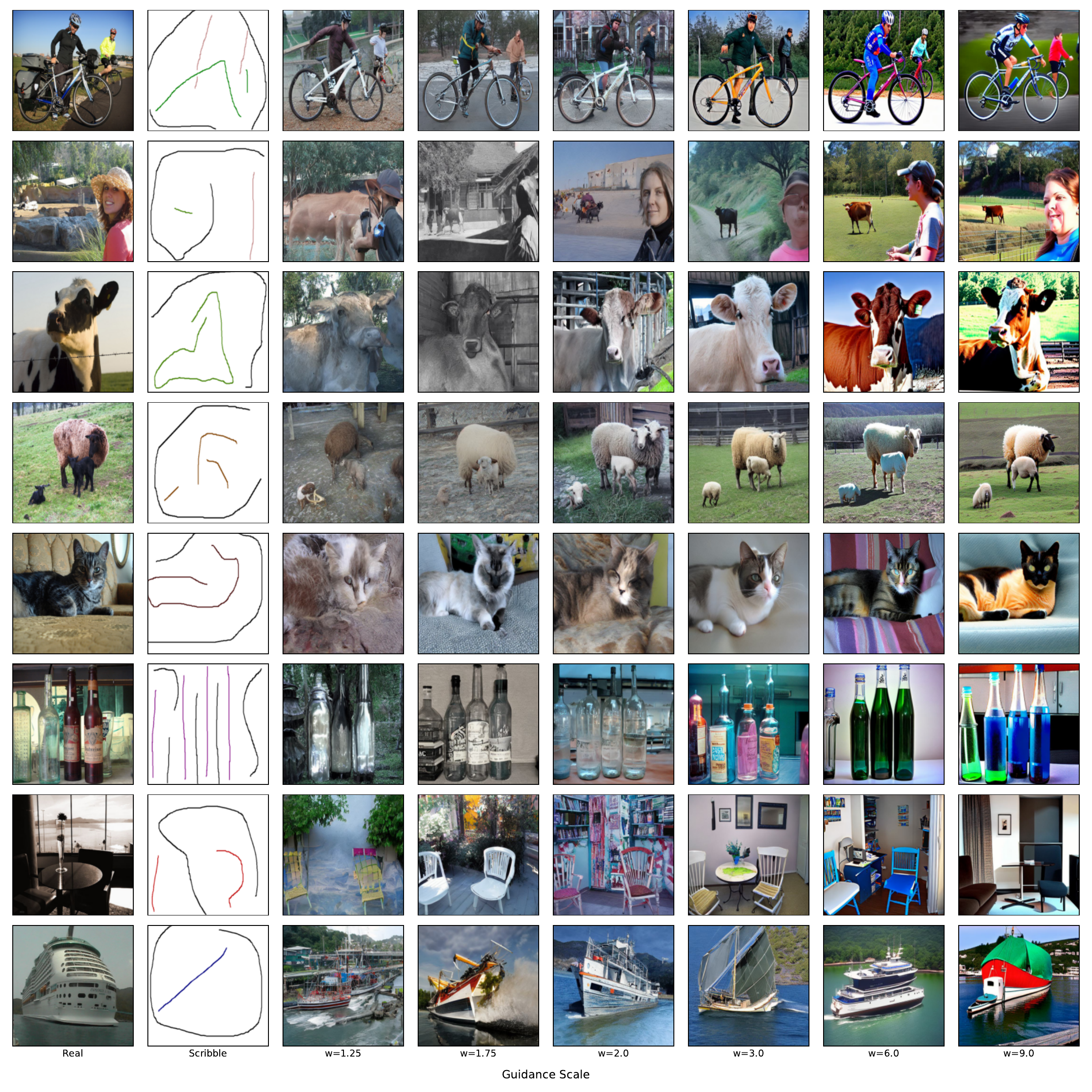}
    \caption{Synthetic training images sampled from a ControlNet model trained on all of scribble-supervised PascalVOC. We vary the guidance scale but keep the encode ratio $\lambda=1.0$ constant to see the effect of the guidance scale on synthesis.}
    \label{fig:guidance-sampling}
\end{figure*}

\begin{figure*}
    \centering
    \includegraphics[width=0.9\textwidth]{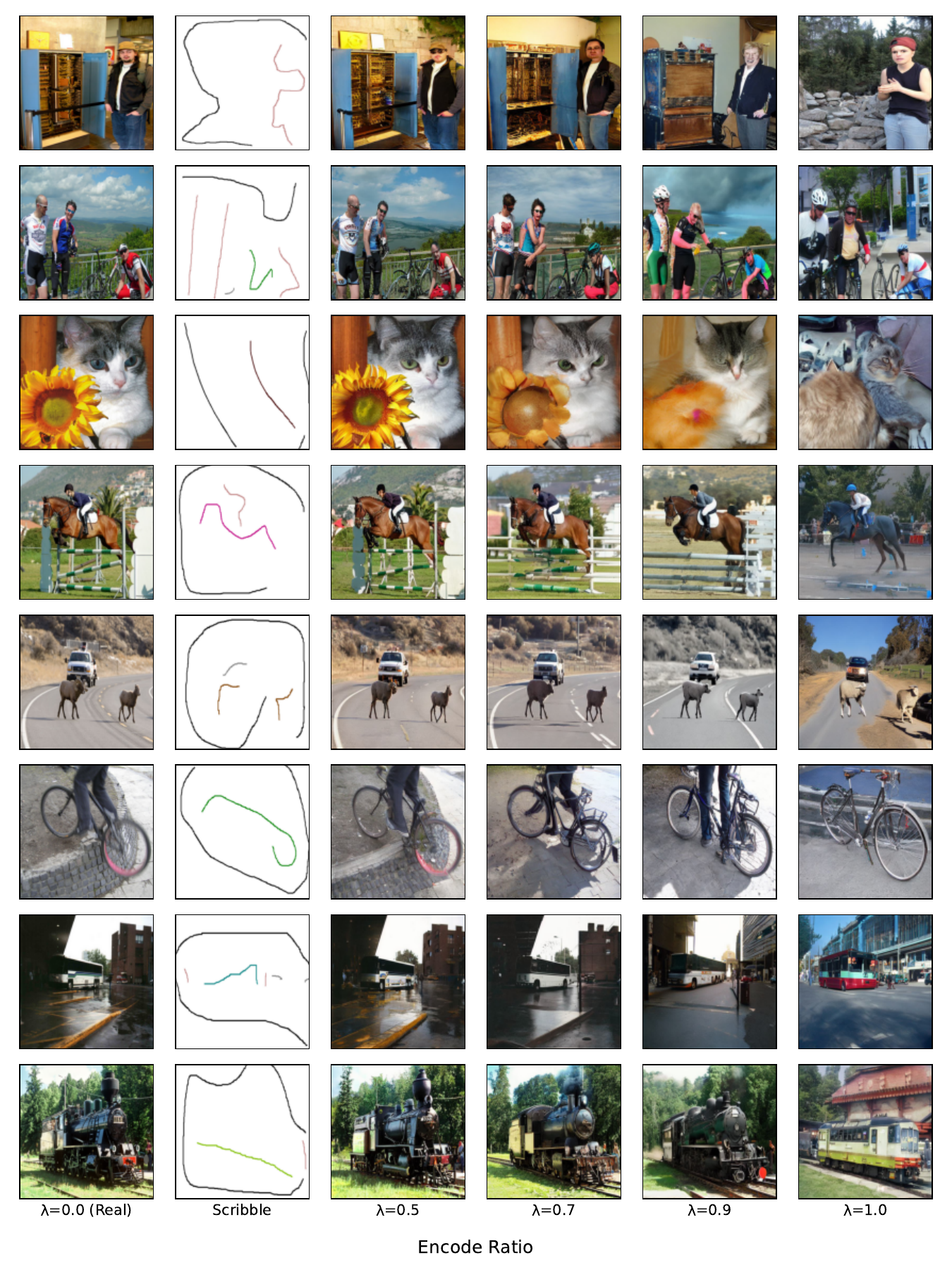}
    \caption{Synthetic training images sampled from a ControlNet model trained on all of scribble-supervised PascalVOC. We vary the encode ratio but keep the guidance scale $w=2.0$ constant to see the effect of the encode ratio on synthesis.}
    \label{fig:encode-sampling}
\end{figure*}

\begin{figure*}
    \centering
    \includegraphics[width=0.9\textwidth]{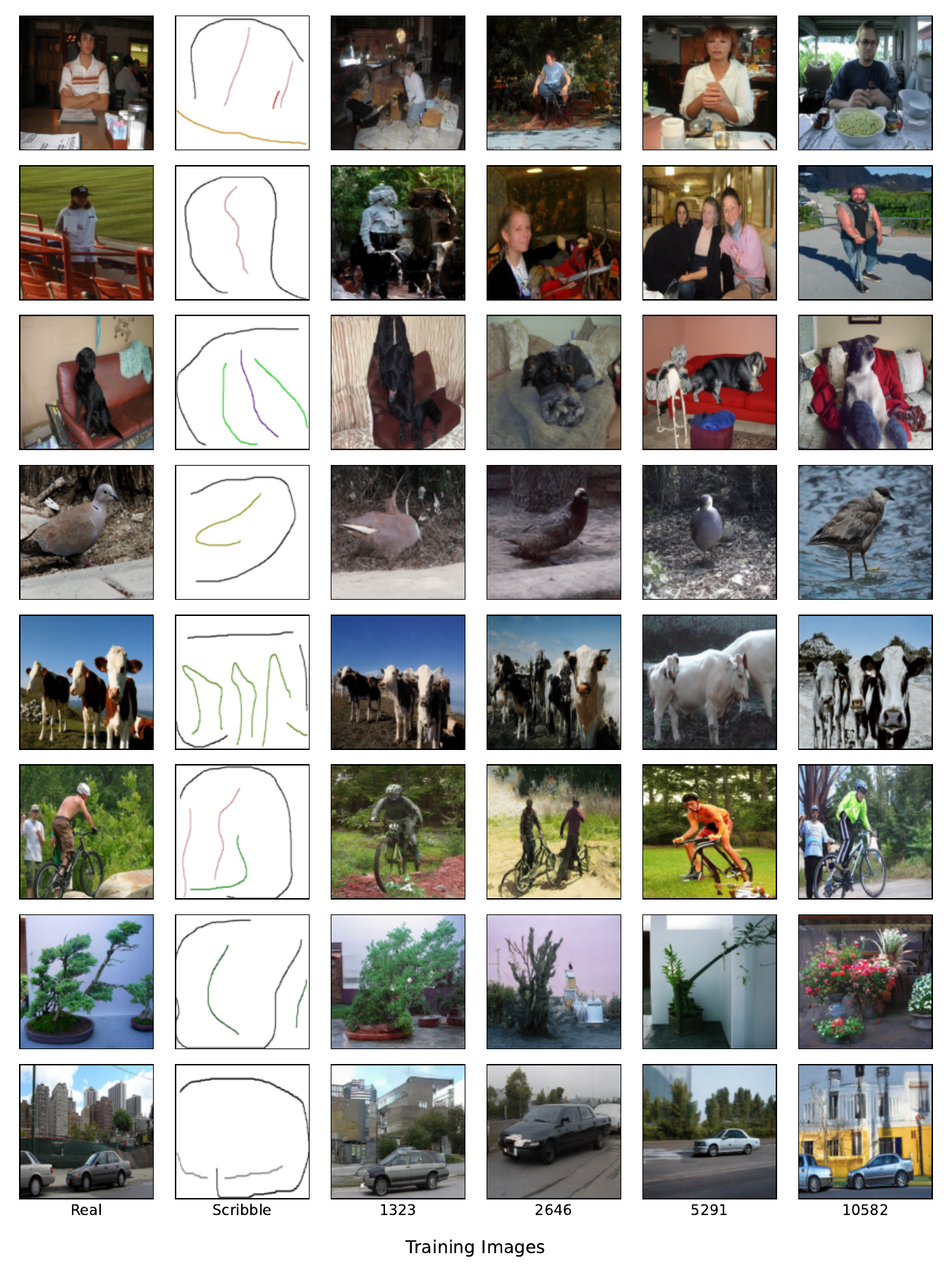}
    \caption{Synthetic training images sampled from a ControlNet model. We vary the number of images on which the ControlNet model is trained to see the impact of the number of training images on synthesis. All images are sampled using guidance scaled $w=2.0$ and encode ratio $\lambda=1.0$.}
    \label{fig:ssl-sampling}
\end{figure*}

\begin{figure*}[h]
\newcommand{\makedemoline}[1]{\includegraphics[width=0.13\linewidth]{images/demos/#1_img.jpg} & \includegraphics[width=0.13\linewidth]{images/demos/#1_baseline.png} & \includegraphics[width=0.13\linewidth]{images/demos/#1_full_resize.png} & 
\includegraphics[width=0.13\linewidth]{images/demos/#1_mixsig04.png} & 
\includegraphics[width=0.13\linewidth]{images/demos/#1_gt.png}
}
    \centering
    \begin{tabular}{ccccc}
        image &  AGMM~\cite{AGMM} & Fully-supervised 
        & AGMM (+syn. images) & ground truth \\
        \makedemoline{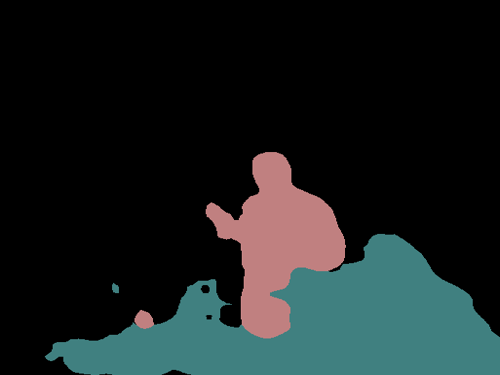} \\
        \makedemoline{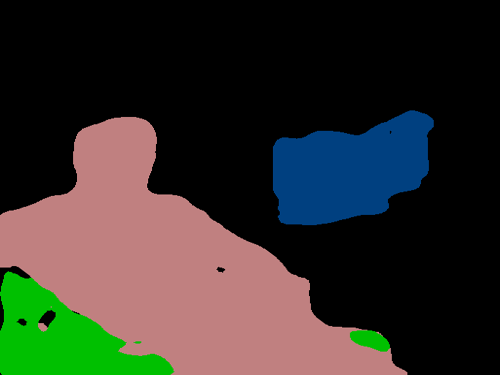} \\
        \makedemoline{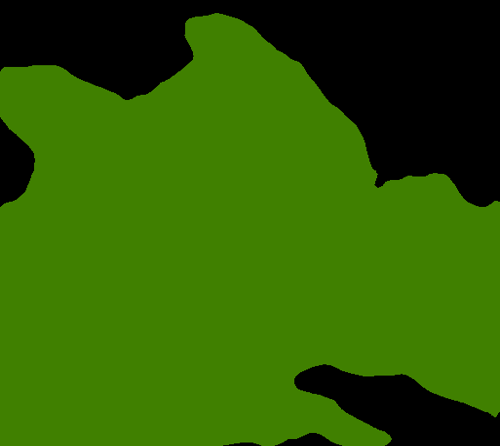} \\
        \makedemoline{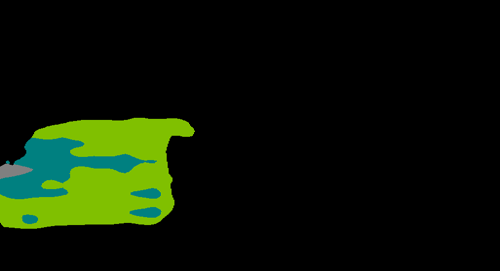} \\
        \makedemoline{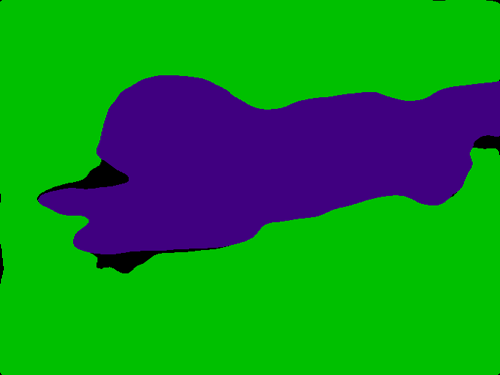} \\
        \makedemoline{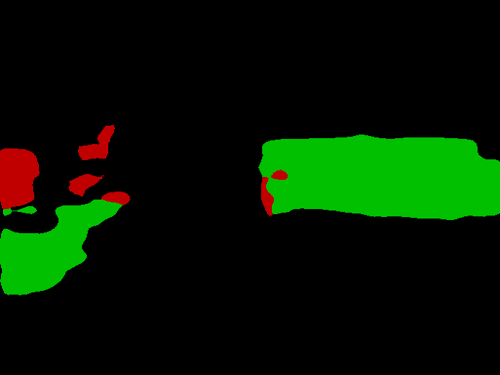} \\
        \makedemoline{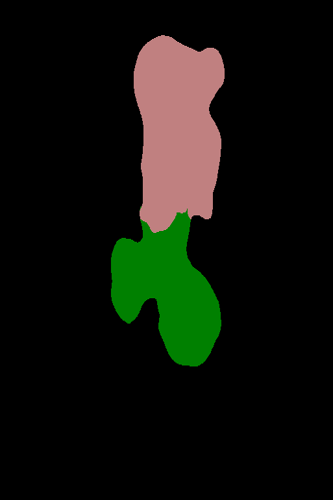} \\
    \end{tabular}
    \caption{Qualitative results on PASCAL dataset. Our generative data augmentation method improves scribble-supervised semantic segmentation method such as AGMM~\cite{AGMM}.}
    \label{fig:more-segmentation}
\end{figure*}

In this section, we include additional qualitative results of our method. 
We provide additional samples of our training data in Fig.~\ref{fig:general-sampling} and samples from previously unseen scribbles from the validation set in Fig.~\ref{fig:validation-sampling}. 
We also provide visualizations of the effect of the guidance scale on synthesis in Fig.~\ref{fig:guidance-sampling} and the effect of the encode ratio in Fig.~\ref{fig:encode-sampling}.
The effect of the number of training images on synthesis is demonstrated in Fig.~\ref{fig:ssl-sampling}.
Finally, we provide further visualizations of segmentation results in Fig.~\ref{fig:more-segmentation}.

\end{document}